\pdfoutput=1

\documentclass[11pt]{article}

\usepackage[final]{ACL2023}

\usepackage{times}
\usepackage{latexsym}
\usepackage{stfloats}

\usepackage[T1]{fontenc}

\usepackage[utf8]{inputenc}

\usepackage{microtype}

\usepackage{inconsolata}
\usepackage{graphicx}
\usepackage{amssymb}
\usepackage{float}
\usepackage{tabularx}
\usepackage{hyperref}

\title{Pruning for Performance: Efficient Idiom and Metaphor Classification in Low-Resource Konkani Using mBERT}

\author{
  Timothy Do,
  Pranav Saran,
  Harshita Poojary,
  Pranav Prabhu,\\
  \textbf{Sean O'Brien},
  \textbf{Vasu Sharma},
  \textbf{Kevin Zhu} \\
  Algoverse AI Research \\
    \texttt{kevin@algoverse.us}
}

\begin{document}
\maketitle
\begin{abstract}
In this paper, we address the persistent challenges that figurative language expressions pose for natural language processing (NLP) systems, particularly in low-resource languages such as Konkani. We present a hybrid model that integrates a pre-trained Multilingual BERT (mBERT) with a bidirectional LSTM and a linear classifier. This architecture is fine-tuned on a newly introduced annotated dataset for metaphor classification, developed as part of this work. To improve the model’s efficiency, we implement a gradient-based attention head pruning strategy. For metaphor classification, the pruned model achieves an accuracy of 78\%. We also applied our pruning approach to expand on an existing idiom classification task, achieving 83\% accuracy. These results demonstrate the effectiveness of attention head pruning for building efficient NLP tools in underrepresented languages.
\end{abstract}

\section{Introduction}
Understanding figurative language is crucial for building NLP systems that can accurately interpret meaning, support effective communication, and preserve cultural nuance \cite{shutova2015design, yang-etal-2025-navajo}. This is especially important for low-resource languages like Konkani \cite{gaonkarDigitization2019}. Improving NLP for Konkani not only advances linguistic research but also contributes to equitable technology access and the safeguarding of linguistic heritage \cite{gaonkarDigitization2019}.
Figurative language expressions like idioms and metaphors are common in Konkani but remain challenging for computational models \cite{shaikh-etal-2024-konidioms}. While such tasks have been explored in major languages, research on Konkani is still emerging \cite{konidioms2024, shaikh-etal-2024-konidioms}. Recent work has introduced the first idiom-annotated corpus and neural models for idiom classification \cite{shaikh-etal-2024-konidioms, shaikh-pawar-2024-identification}, but these efforts are limited. They focus solely on idioms, neglect metaphor classification, and do not consider model efficiency improvements.

\begin{figure}[t]
  \centering
  \includegraphics[width=0.35\textwidth]{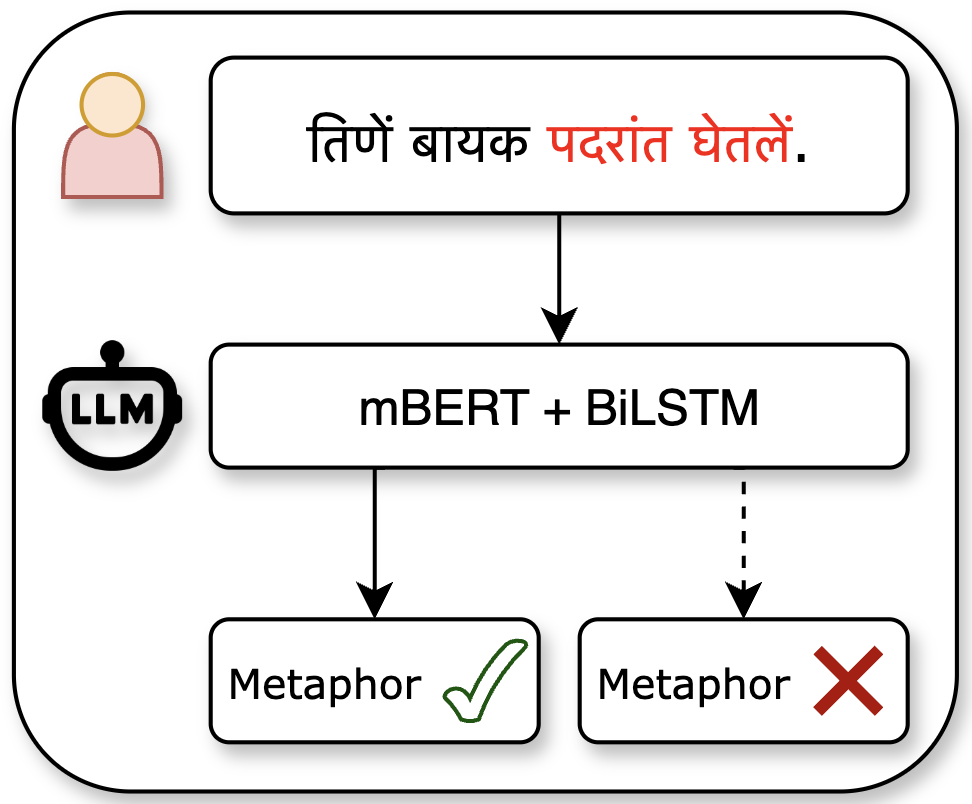}
  \caption{Processing of Konkani metaphorical expressions using mBERT+BiLSTM. The phrase highlighted in red is analyzed for metaphorical content, with contrasting classification outcomes shown.}
  \label{fig:representative-figure}
\end{figure}

We present a hybrid model that integrates a pre-trained Multilingual BERT (mBERT) \cite{devlin-etal-2019-bert} with a bidirectional LSTM and a linear classifier, as shown in Figure \ref{fig:representative-figure}. This architecture is fine-tuned on an adapted version of the Konidioms corpus \cite{shaikh-etal-2024-konidioms}, which we extend to include metaphor annotations. To improve efficiency, we apply gradient-based attention head pruning. Our results show that pruning significantly reduces model complexity, with one experiment maintaining performance and the other showing a small decline. These findings demonstrate the effectiveness of pruning for building efficient NLP models in low-resource settings.

\section{Related Work}
Research on low-resource languages has underscored challenges such as limited annotated data, script diversity, and dialectal variation \cite{rajan2020survey, yang2025nushurescue, arxiv2024lowresource, gaonkarDigitization2019}. Konkani reflects these issues through its use of multiple scripts, dialectal fragmentation, and a shrinking speaker population. Prior work has addressed tasks like text summarization using a small folk tale dataset and language-independent features with pre-trained embeddings \cite{automatictextsummarizationsharma}, but figurative language remains largely unexplored.

\citet{shaikh-etal-2024-konidioms} introduced the first idiom-annotated corpus of 6,520 Devanagari-script sentences, and \citet{shaikh-pawar-2024-identification} developed a neural classifier. \citet{yayavaram-etal-2024-bert} further improved idiom classification using a BERT-based model with custom loss functions. To improve model efficiency, especially in low-resource settings, recent studies have explored pruning redundant attention heads. \citet{Feng_2018} showed that gradients can assess feature importance, and \citet{Ma_2021} extended this to cross-lingual attention head pruning.

Building on this, we adopt a gradient-based attention head pruning strategy, which identifies and removes less important Transformer attention heads by analyzing gradient magnitudes during backpropagation \citep{michel2019sixteenheadsreallybetter}. This approach not only reduces model complexity and memory usage, critical in extreme low-resource environments like Konkani, but also enhances interpretability by revealing which heads capture task-relevant figurative patterns. To our knowledge, this is the first application of Transformer attention head pruning for any NLP task in Konkani, providing both practical efficiency and novel linguistic insights for metaphor classification.

\subsection{Konkani Language}
Konkani is an Indo-Aryan language spoken along India’s western coast, classified within the Southern Indo-Aryan Outer Languages branch alongside Marathi (Figure~\ref{fig:language_tree}) \cite{rajan2020survey, gaonkarDigitization2019}. With approximately 2.5 million speakers \cite{Britannica_Konkani} concentrated in the coastal regions of western India (Figure~\ref{fig:geographic_map}), the language faces endangerment due to dialectal fragmentation and limited digital resources, despite ongoing corpus development efforts \cite{gaonkarDigitization2019}. This precarious situation underscores the urgency of preserving Konkani not only as a medium of communication but also as a vessel of cultural identity, as echoed by native speakers' reflections and personal narratives (Appendix~\ref{sec:appendix_b}).

\begin{table}[t]
  \raggedleft
  \resizebox{0.48\textwidth}{!}{%
    \begin{tabular}{|l|p{0.4\textwidth}|}
      \hline
      \textbf{Id}                  & Sentence instance identifier                                   \\\hline
      \textbf{Expression}          & The expression in Konkani                      \\\hline
      \textbf{Sentence}            & Konkani sentence with the expression      
      \\\hline
      \textbf{Idiom}               & Identification tag for Idioms (Yes/No)                         \\\hline
      \textbf{Metaphor}            & Identification tag for Metaphors (Yes/No)                      \\\hline
      \textbf{Split}               & Data split assignment (train or test)                          \\\hline
    \end{tabular}%
  }
  \caption{Data schema for modified Konidioms Corpus.}
  \label{tab:dataset_columns}
\end{table}

\section{Metaphor Classification}
To our knowledge, \textbf{this is the first work to introduce a metaphor-annotated dataset for the Konkani language in NLP}. We extend the Konidioms Corpus \cite{shaikh-etal-2024-konidioms} by manually labeling 500 sentences with binary metaphor annotations. All labels were verified by three native Konkani speakers for linguistic accuracy. Table~\ref{tab:dataset_columns} shows the structure of an annotated entry.

For evaluation, we curated a balanced dataset of 200 sentences (50\% metaphorical, 50\% literal) and split it into an 80/20 train-test set to avoid class imbalance and ensure consistent training. Hyperparameters are detailed in Appendix~\ref{sec:appendix_d}.

Building on prior work in attention head pruning and transformers, we propose the first application of this technique to metaphor classification in Konkani. We also apply it to idiom classification, previously explored in earlier work, to highlight its broader relevance. Figure~\ref{fig:pipeline} in Appendix~\ref{sec:appendix_c} provides a high-level overview of our methodology.

\section{Results}
The comparison between original and pruned models reveals differential impacts across the two figurative language classification tasks, as shown in Table~\ref{tab:eval-combined-classification}. For idiom classification, pruning resulted in remarkably stable performance, with minor but consistent gains. Accuracy increased from 0.82 to 0.83, and recall improved from 0.89 to 0.91, while the F1-score remained stable at 0.88. This stability extended to macro and weighted averages across all metrics, with changes typically within 0.01–0.02 points. These results indicate that the pruned attention heads contributed minimally to idiom detection capabilities, highlighting the model’s robustness to compression.

In contrast, metaphor classification exhibited greater sensitivity to pruning, with more pronounced declines across all evaluation metrics. Accuracy dropped from 0.88 to 0.78, and both precision and recall declined substantially, leading to a lower F1-score. Macro and weighted average metrics each fell by approximately 0.10 points. This performance drop reflects the model’s reliance on a broader and more distributed set of attention heads for metaphor detection. These results underscore that while pruning can improve or maintain performance for certain tasks like idiom classification, it may significantly degrade performance in others, reinforcing the importance of task-specific pruning strategies.

\begin{figure*}[t]
  \centering
  \includegraphics[width=\textwidth]{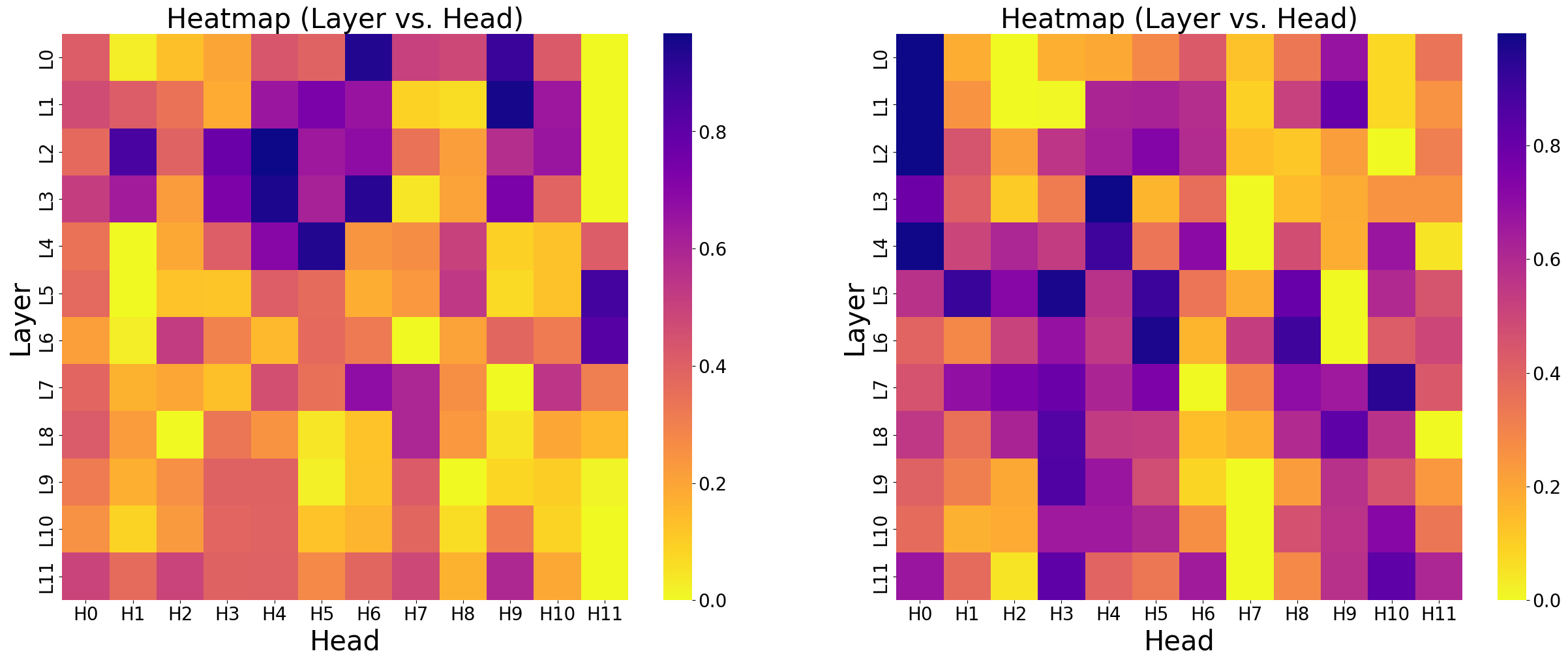}
  \caption{Heatmaps showing attention head importance scores across layers for idiom (left) and metaphor (right) classification. Idiom classification shows higher importance values in earlier layers compared to later ones, while metaphor classification exhibits a wider spread of higher importance values across the layers.}
  \label{fig:heatmap}
\end{figure*}

\section{Attention Head Analysis}
We prune attention heads in the mBERT component of the mBERT+BiLSTM model using a gradient-based importance metric \cite{michel2019sixteen}. This metric quantifies each head's contribution by calculating the expected sensitivity of the model loss to the head's removal, expressed as $I_h = \mathbb{E}_{(x,y)\sim D} \left| \frac{\partial L}{\partial \mathbf{h}^{(h)}} \right|$, where $I_h$ is the importance score for head $h$, $(x,y)$ represents input-output pairs from dataset $D$, $L$ is the loss, and $\mathbf{h}^{(h)}$ is the output of attention head $h$. For each of the 144 heads (12 layers $\times$ 12 heads), we compute the average absolute gradient of the loss with respect to the head's output. Heads with scores of zero were pruned post hoc, with no changes to the BiLSTM.

For both idiom and metaphor classification tasks, we pruned all attention heads that had an importance score of zero, resulting in 132 of 144 heads being retained for both tasks. The attention head maps can be seen in Figure \ref{fig:heatmap}. By eliminating these attention heads with zero importance scores across both tasks, we create two pruned variants of the original model. These pruned models are evaluated and compared against the baseline. These results are presented in Table \ref{tab:eval-combined-classification}.

\subsection{Head-Level Performance}
Figure~\ref{fig:heatmap} visualizes the distribution of attention head importance for both idiom and metaphor classification tasks. For idiom classification, importance tends to cluster in the lower layers (\textbf{L0--L6}), with particularly prominent heads such as \textbf{L0-H6} and \textbf{L1-H9} standing out as key contributors. These heads likely encode lexical or syntactic patterns crucial for identifying idiomatic usage. In contrast, metaphor classification exhibits a more diffuse pattern of importance, with salient heads scattered across all layers. This broader distribution suggests that metaphor detection may require integrating cues from multiple linguistic levels. Despite some variation, both tasks reveal consistent retention of highly informative heads, supporting the effectiveness of selective pruning in reducing model complexity without compromising performance.

The contrasting patterns observed in the two classification tasks, suggests fundamental differences in how these separate linguistic classification problems are processed within the transformer's attention mechanism. Full detailed heatmaps for idiom and metaphor classification can be found in Appendix~\ref{sec:appendix_c} (Figure \ref{fig:detailed_heatmap_idiom} and Figure \ref{fig:detailed_heatmap_metaphor} respectively).

\section{Discussion}

\begin{table*}[t]
  \centering
  \resizebox{\textwidth}{!}{%
    \begin{tabular}{lcccccc}
      \hline
      \textbf{Metric} & \multicolumn{2}{c}{\textbf{Idiom Classification}} & & \multicolumn{2}{c}{\textbf{Metaphor Classification}} \\
      \cline{2-3} \cline{5-6}
      & \textbf{Original Model} & \textbf{Pruned Model} & & \textbf{Original Model} & \textbf{Pruned Model} \\
      \hline
      Precision  & 0.87 & 0.86 & & 1.00 & 0.87 \\
      Recall     & 0.89 & 0.91 & & 0.75 & 0.65 \\
      F1-Score   & 0.88 & 0.88 & & 0.86 & 0.74 \\
      Accuracy   & 0.82 & 0.83 & & 0.88 & 0.78 \\
      Macro Avg Precision & 0.78 & 0.79 & & 0.90 & 0.79 \\
      Macro Avg Recall    & 0.77 & 0.77 & & 0.88 & 0.78 \\
      Weighted Avg Precision & 0.82 & 0.82 & & 0.90 & 0.79 \\
      Weighted Avg Recall    & 0.82 & 0.83 & & 0.88 & 0.78 \\
      \hline
    \end{tabular}%
  }
  \caption{\label{tab:eval-combined-classification}
    Comparison of original and pruned mBERT+BiLSTM models on idiom and metaphor classification. Idiom performance remains stable post-pruning, while metaphor classification shows metric drops, reflecting its reliance on a broader set of attention heads and the need for task-specific pruning strategies.
  }
\end{table*}

The heatmaps in Figure~\ref{fig:heatmap} reveal why pruning affects idiom and metaphor classification differently. Idiom classification shows higher importance in early layers, allowing redundancy that preserves performance even after pruning. In contrast, metaphor classification has a more distributed pattern with mid-layer importance, making it more sensitive to head removal.

This structural difference aligns with our experimental results: idiom  classification remained stable post-pruning, while \textbf{metaphor classification saw consistent performance drops} across all metrics. This suggests metaphor detection depends on a more intricate, interconnected attention structure that pruning disrupts.

We chose the mBERT+BiLSTM architecture based on both empirical results and the constraints of low-resource settings. Prior research shows BiLSTMs can outperform BERT by over 16\% when trained on just 25\% of the data \cite{ezencan2020comparisonlstmbertsmall}, though this gap narrows with larger datasets. Given the limited annotated data for endangered languages, our goal was to maximize interpretability, efficiency, and cross-lingual transfer. The BiLSTM layer complements mBERT by capturing sequential context, enhancing robustness even under pruning. Our ablation results validated this: after pruning 8.33\% of parameters, our selected model outperformed all pruned baselines (Table~\ref{tab:ablation-results}, Appendix~\ref{sec:appendix_d}).

These findings have important implications for pruning in low-resource NLP. They show that pruning must be task-specific. For idiom classification, pruning is effective and efficient, but for metaphor detection, aggressive pruning undermines performance. A one-size-fits-all pruning strategy is therefore unsuitable for figurative language tasks with different attention head distributions.

Future work should explore adaptive pruning methods that tailor compression to each task’s architectural needs. Varying pruning thresholds could further reveal how performance degrades under different constraints. Additionally, expanding the dataset would help reduce overfitting and improve generalization, supporting the development of efficient and task-specific pruning strategies for figurative language understanding in multilingual and low-resource environments.

\section{Conclusion}
We introduce the first metaphor-annotated dataset for Konkani and apply a unified framework for idiom and metaphor classification in a low-resource setting. By extending the Konidioms corpus and fine-tuning a hybrid mBERT+BiLSTM model, we establish strong baselines for figurative language understanding. Gradient-based attention head pruning reveals structural differences: idioms rely on localized, lower-layer heads, while metaphors engage a more diffuse attention profile. As a result, idiom classification remains robust under pruning, whereas metaphor performance is more sensitive to head removal. Our work advances interpretable NLP for underrepresented languages. We release our dataset and pruning framework to support future research in figurative language modeling, model compression, and multilingual generalization.

\section*{Limitations}
This study is limited by several key factors. Although the metaphor classification dataset includes 500 newly annotated data points, our experiment utilized only 200 balanced sentences, which limits the generalizability of our results and highlights the need for broader evaluation in future work. Although we verified annotations with three native Konkani speakers, the small number of validators introduces potential subjective bias in the labeling process. The corpus itself may not capture the full range of figurative expressions or dialectal variations present in Konkani, affecting model performance across different speaker communities. Our pruning approach, while effective for our experiments, employed fixed thresholds that may not transfer optimally to other tasks or datasets. Finally, evaluation on a single test split necessitates further validation with more diverse data to confirm the robustness of our findings across different contexts.

\section*{Ethics Statement}
Our research addresses the technological gap between high and low-resource languages while recognizing the ethical responsibilities inherent in working with Konkani, an endangered language \cite{bird2020decolonising}. We engaged native speakers throughout the annotation and verification process to ensure linguistic accuracy and cultural sensitivity. This work contributes to preserving Konkani's cultural heritage by documenting and enabling computational processing of its figurative expressions. The resources we have developed are intended to serve both the Konkani-speaking community and researchers working on low-resource language technologies \cite{bird2024must}. We have maintained transparency about our limitations to prevent misrepresentation of capabilities, and our pruning approach specifically addresses accessibility in resource-constrained environments. By balancing our dataset and committing to continued community engagement, we aim to support linguistic diversity and ensure all languages receive technological support that preserves their unique characteristics in digital spaces. In the spirit of transparency, our code is made publicly available in an anonymous repository at \href{https://anonymous.4open.science/r/KonkaniNLP}{https://anonymous.4open.science/r/KonkaniNLP}.

\bibliography{anthology,custom}
\bibliographystyle{acl_natbib}

\clearpage
\appendix

\section{Appendix A}
\label{sec:appendix_a}
\begin{figure}[!ht]
    \centering
    \includegraphics[width=0.8\textwidth]{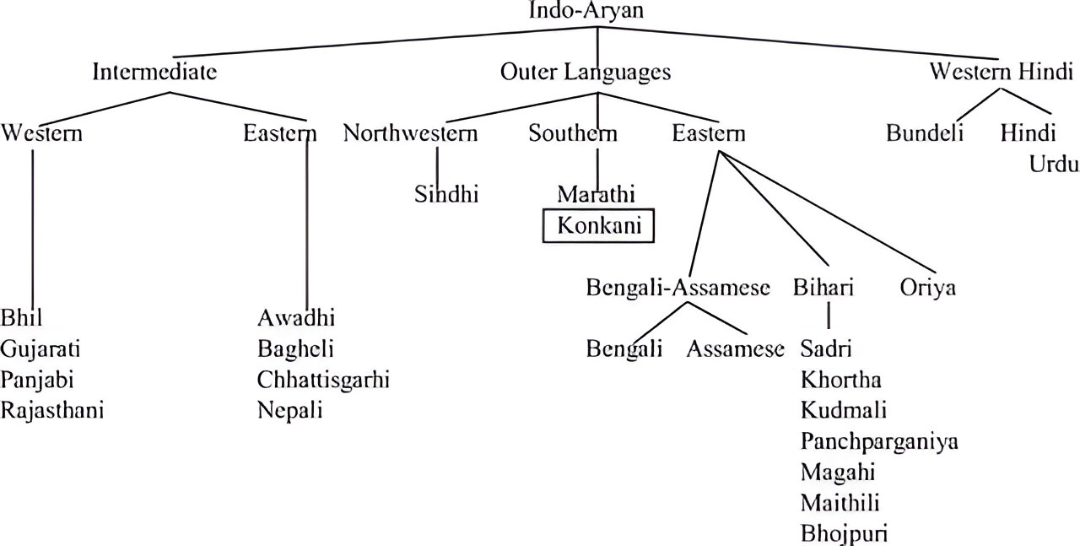}
    \caption{Linguistic tree showing Konkani's classification as a Southern language within the Indo-Aryan Outer Languages branch, alongside Marathi and distinct from other major Indo-Aryan language groups.}
    \label{fig:language_tree}
\end{figure}

\vspace{1cm} 

\begin{figure}[!ht]
    \centering
    \includegraphics[width=0.8\textwidth]{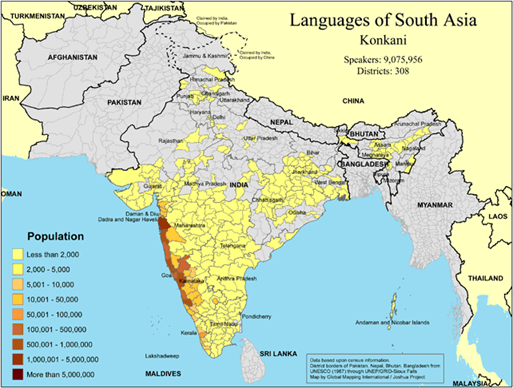}
    \caption{Geographic distribution of Konkani speakers across South Asia, concentrated along India's western coastal regions. As of 2018, approximately 9 million speakers were recorded across 308 districts. Source: \protect\url{https://www.missioninfobank.org/mib/index.php?main_page=product_info&products_id=6368}}
    \label{fig:geographic_map}
\end{figure}

\clearpage
\section{Appendix B}
\label{sec:appendix_b}

\subsection{Perspectives from a Native Konkani Speaker}
As part of this work, we solicited reflections from a native Konkani speaker regarding the digital and computational underrepresentation of the language. The following excerpt is shared with permission and reflects the perspective of a native speaker from Goa:

\begin{quote}
“As a native Konkani speaker from Goa, I find it deeply concerning that Konkani remains a low-resource language in the digital world today. Although spoken by hundreds of thousands and recognized as one of India's official languages, Konkani lacks the technological and academic investment that the more dominant languages receive. This underrepresentation threatens the long-term vitality of our language, culture, and identity.

Languages like Konkani are not just modes of communication, they are carriers of unique histories, worldviews, and traditions. When they are ignored by major platforms, AI models, and digital tools, it sends the message that these voices matter less. But they do matter.

I believe that it is our responsibility as speakers, researchers, and technologists to change that. Supporting Konkani through language research, resource development, and digital inclusion is not just about preserving a language. It's about empowering a community.”
\end{quote}

\noindent\hfill\textit{— Native Konkani speaker from Goa}

\vfill \break

\subsection{In Memory of a Monolingual Konkani Speaker}

This project is motivated in part by the memory of a monolingual speaker of Konkani whose life, conversations, and cultural expressions were deeply rooted in the language. His use of idioms and metaphors exemplified the richness and complexity of Konkani, elements that are often difficult to preserve or translate into other languages.

His recent passing highlights the urgency of documenting and understanding low resource languages like Konkani, not only from a linguistic perspective, but also as a means of preserving cultural and emotional heritage. This research, particularly its focus on idiomatic and metaphorical structures, reflects a commitment to honoring such speakers and the languages they embody.

We hope that advancements in AI models capable of capturing linguistic nuance may one day help reflect not just the syntax, but the soul of languages like Konkani.

\clearpage
\section{Appendix C}
\label{sec:appendix_c}

\begin{figure}[h!]
\centering
\sbox0{\includegraphics[width=0.9\textwidth]{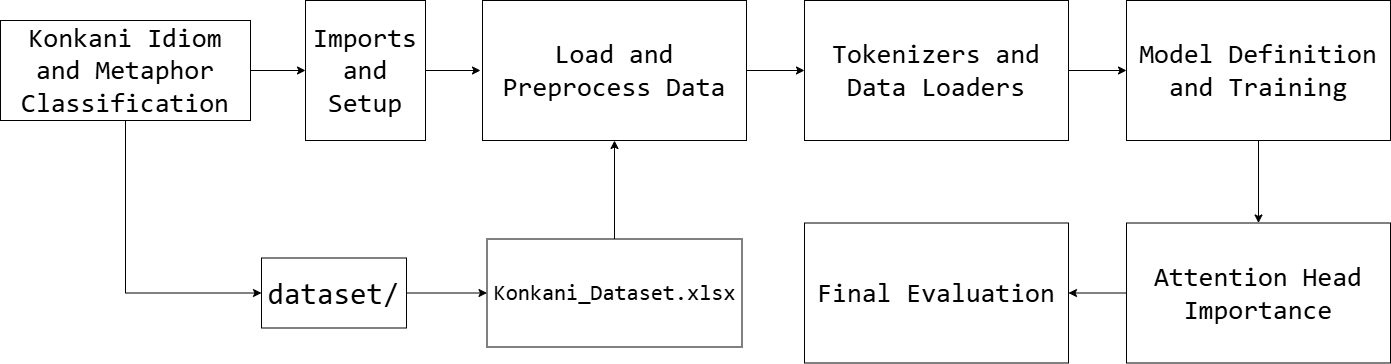}}
\begin{minipage}{\wd0}
\usebox0
\caption{Flowchart outlining our experimental pipeline.}
\label{fig:pipeline}
\end{minipage}
\end{figure}

\begin{figure}[h!]
\centering
\sbox0{\includegraphics[width=0.85\textwidth]{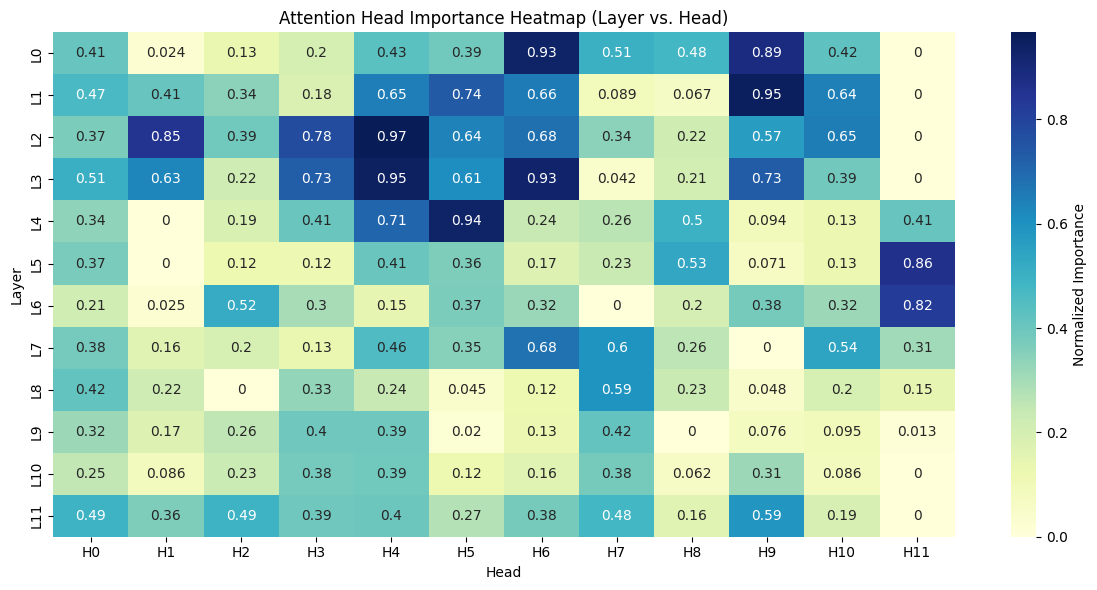}}
\begin{minipage}{\wd0}
\usebox0
\caption{Heatmap visualization of attention head importance across model layers for idiom classification, with numerical decimal values displayed to facilitate detailed quantitative analysis.}
\label{fig:detailed_heatmap_idiom}
\end{minipage}
\end{figure}

\begin{figure}[h!]
\centering
\sbox0{\includegraphics[width=0.85\textwidth]{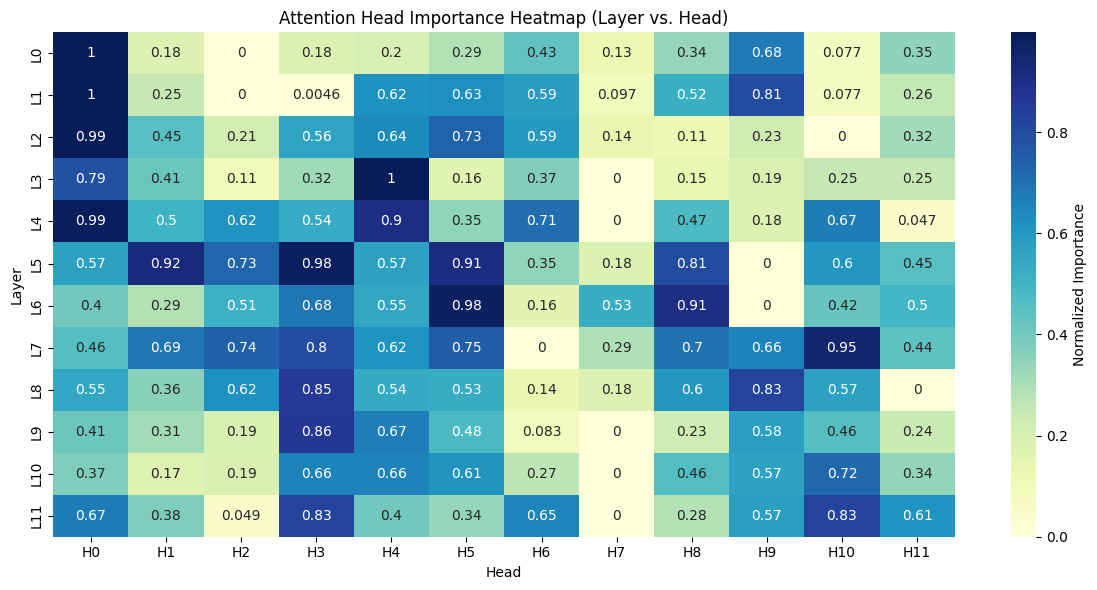}}
\begin{minipage}{\wd0}
\usebox0
\caption{Heatmap visualization of attention head importance across model layers for metaphor classification, with numerical decimal values displayed to facilitate detailed quantitative analysis.}
\label{fig:detailed_heatmap_metaphor}
\end{minipage}
\end{figure}

\clearpage
\section{Appendix D}
\label{sec:appendix_d}

\begin{table}[ht]
\centering
\resizebox{\textwidth}{!}{
\begin{tabular}{|l|l|c|c|c|c|c|}
\hline
\textbf{Model} & \textbf{Task} & \textbf{Prune \%} & \textbf{Acc. (Orig.)} & \textbf{Acc. (Pruned)} & \textbf{F1 (Orig.)} & \textbf{F1 (Pruned)} \\
\hline
mBERT & Idiom & 8.33 & 0.81 & 0.72 & 0.77 & 0.42 \\
mBERT & Metaphor & 8.33 & 0.93 & 0.50 & 0.92 & 0.33 \\
\textbf{mBERT + BiLSTM} & \textbf{Idiom} & \textbf{8.33} & \textbf{0.82} & \textbf{0.83} & \textbf{0.78} & \textbf{0.78} \\
\textbf{mBERT + BiLSTM} & \textbf{Metaphor} & \textbf{8.33} & \textbf{0.88} & \textbf{0.78} & \textbf{0.87} & \textbf{0.77} \\
IndicBERT & Idiom & 8.33 & 0.87 & 0.72 & 0.84 & 0.42 \\
IndicBERT & Metaphor & 8.33 & 0.90 & 0.53 & 0.90 & 0.39 \\
XLM-R + BiLSTM + Attn & Idiom & 8.33 & 0.78 & 0.28 & 0.74 & 0.22 \\
XLM-R + BiLSTM + Attn & Metaphor & 8.33 & 0.78 & 0.50 & 0.77 & 0.33 \\
\hline
\end{tabular}
}
\caption{\parbox{\textwidth}{ Ablation results before and after pruning across different models and tasks.}}

\label{tab:ablation-results}
\end{table}

We fine-tuned a multilingual BERT (mBERT) model~\cite{devlin-etal-2019-bert} combined with a two-layer BiLSTM (128 hidden units) using standard fine-tuning settings. Training was performed with the AdamW optimizer, a learning rate of $2 \times 10^{-5}$, batch size of 16, and a maximum input length of 128 tokens. A sigmoid-activated linear layer followed the BiLSTM to produce the final output. The model was trained using binary cross-entropy loss for up to 20 epochs, with early stopping applied if validation loss did not improve for 10 consecutive epochs. The best-performing model, selected based on minimum validation loss, balances computational efficiency and representational capacity for detecting idioms and metaphors.

\end{document}